\documentclass[11pt]{article}
\usepackage{eacl2017}
\usepackage{times}
\usepackage{caption}
\usepackage{url}
\usepackage{latexsym}
\usepackage{amsmath}
\usepackage{ragged2e}

\usepackage{lipsum}

\newcommand\blfootnote[1]{%
  \begingroup
  \renewcommand\thefootnote{}\footnote{#1}%
  \addtocounter{footnote}{-1}%
  \endgroup
}

\aclfinalcopy 
\def\aclpaperid{***} 

\title{User Bias Removal in Review Score Prediction}

\author{Rahul Wadbude \\ IIT Kanpur  \\{\tt warahul'}
        \And
        Vivek Gupta \\ Microsoft Research \\ {\tt t-vigu*}
        \And
        Dheeraj Mekala \\ IIT Kanpur \\{\tt dheerajm'}
        \And
        Harish Karnick \\ IIT Kanpur \\{ \tt hk'} 
        }

\date{}

\begin{document}
\maketitle

\begin{abstract}
\blfootnote{'@iitk.ac.in, *@microsoft.com}
Review score prediction of text reviews has recently gained a lot of attention in recommendation systems. A major problem in models for review score prediction is the presence of noise due to user-bias in review scores. We propose two simple statistical methods to remove such noise and improve review score prediction. Compared to other methods that use multiple classifiers, one for each user, our model uses a single global classifier to predict review scores. We empirically evaluate our methods on  two major categories (\textit{Electronics} and \textit{Movies and TV}) of the  SNAP published Amazon e-Commerce Reviews data-set and Amazon \textit{Fine Food} reviews data-set. We obtain improved review score prediction for three commonly used text feature representations. 

\end{abstract}

\section{Introduction}

\subsection{User Bias Problem}
\begin{figure}
\centering
\includegraphics[scale=0.26]{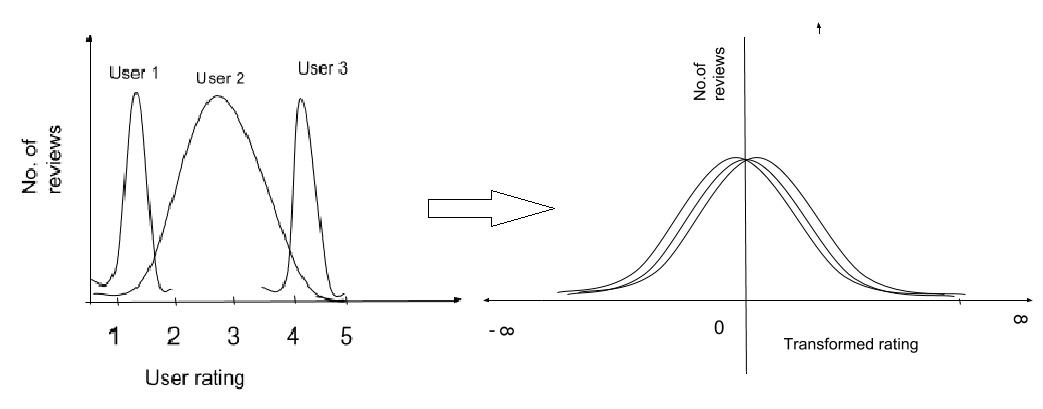}
\caption{User Bias Problem}
\label{biasfigure}
\end{figure}

Different users generally do not rate food or e-commerce products on the same scale. Every user has his/her own preferred scale of rating a product. Some users are generous and rate an item as 4 or 5 (out of {1, 2, 3, 4, 5}), thus introducing a positive bias. At the other extreme, some users may give 1 or 2 (out of {1, 2, 3, 4, 5}), thus introducing a negative bias in the scores. These preferred rating choices of particular users make it difficult to learn a general model for review score prediction. Thus, user-bias removal is a problem that must be handled for accurate review score prediction. Figure \ref{biasfigure}(L) shows score distribution of three users with negative(user 1), neutral(user 2) and positive(user 3) biases. Figure \ref{biasfigure}(R) shows the score distribution of three users after bias removal. user-bias removal methods try to avoid spurious good or bad reviews given by users who always tend to upvote or downvote irrespective of the item quality.

\begin{figure}
\centering
\includegraphics[scale=0.25]{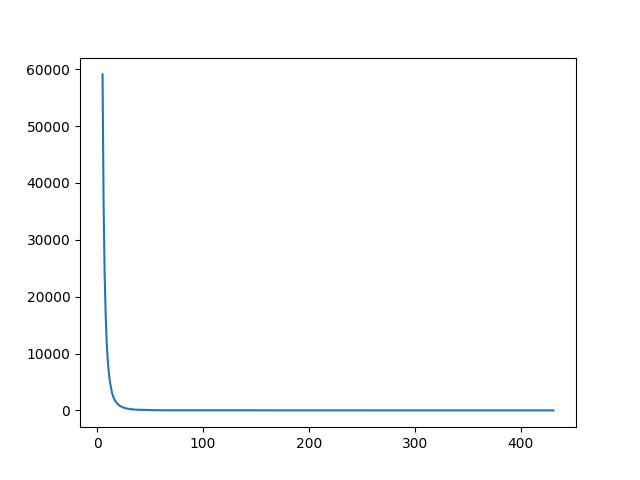}
\caption{Shows heavy tail distribution for Amazon Electronic dataset, where x-axis is number of review rated and y-axis number so users}
\label{tailfigure}
\end{figure}

Past works that perform user-bias modelling and use review text for score prediction focus on building user-specific regressors that predict ratings for a single user only \cite{seroussi2010collaborative} \cite{li2011incorporating}. Most users don't review that often as indicated in Figure \ref{tailfigure}, leading to insufficient training instances. Furthermore, these models are computationally expensive while training and evaluation and have huge storage requirements. Matrix factorization techniques such as those mentioned in \cite{koren2009matrix} and \cite{ricci2011introduction} model user-bias as a matrix factorization problem and rely only on collaborative filtering to predict review scores. They do not make use of the review text for prediction of scores.
In our approach, we build two simple yet universal statistical models (UBR-I and UBR-II) that estimate user-bias for all users. We then learn a single regressor on review text and unbiased review scores to predict unknown scores given corresponding reviews. The main contributions of our work are:
\begin{enumerate}
    \setlength\itemsep{0.5em}
    \item A simple, global user-bias model and a single linear regressor to predict review scores from review text.
    \item Model is computationally efficient with reduced sample, time and space requirements during training and testing.
\end{enumerate}

In Section 2, we present the UBR-I and UBR-II techniques in detail, followed by thorough experiments in Section 3 and a discussion of relevant previous work in Section 4.

\section{Our User-Bias Removal (UBR) Methods}
This section explains the proposed model for rectifying user-bias to improve score prediction.
\label{biasremoval}
 We remove user-bias in scores corresponding to each user ($u_{i}$) by learning a statistical mapping from a user specific scale to a general scale common to all users. We propose two methods to learn such a mapping, as described in detail below.

\subsection{User-Bias Removal-I (UBR-I)}
 We develop a user specific statistical mapping for user-bias removal, by normalizing each review score with respect to the mean and standard deviation of all products rated by that user. During prediction we use the same user specific mean and standard deviation (statistical mapping) to revert back to the original scale. 

Let $R(u_{i},p_{j})$ represent the review score of user $u_{i}$ for product $p_{j}$. We calculate the normalized score $NR(u_{i},p_{j})$ for training, and predict score $PR(u_{k},p_{m})$ for new review of user $u_{k}$ for product $p_{m}$ during prediction as follows:-

\begin{enumerate}
    \item For each user, calculate and store the mean $R_{\mu}(u_{i})$ of all  scores given by user $u_{i}$. 
    \begin{equation*}
    R_{\mu}(u_{i}) = \frac{1}{N_{u_{i}}} \sum_{j=1}^{N_{u_{i}}} {R(u_{i},p_{j})}
    \end{equation*}
     Here, $N_{u_{i}}$ represents the number of products reviewed by user $u_{i}$.
     \item Similarly, for every user calculate and store standard deviation $R_{\sigma}(u_{i})$ of all the scores given by the user $u_{i}$.     \begin{equation*}
        R_{\sigma}(u_{i}) = \sqrt{\frac{1}{N_{u_{i}}} \sum_{j=1}^{N_{u_{i}}} (R(u_{i},p_{j}) - R_{\mu}(u_{i}))^2}
     \end{equation*} 
    \item For every review score, calculate the Normalised user-bias removed score $NR(u_{i},p_{j})$ as follows :
     \begin{equation*}
       NR(u_{i},p_{j}) = \frac{1}{R_{\sigma}(u_{i})} (R(u_{i},p_{j}) - R_{\mu}(u_{i}))
     \end{equation*}
     Here, $NR(u_{i},p_{j})$ represents the normalised score (after user-bias removal) for user $u_{i}$ and product $p_{j}$. In the trivial case when $R_{\sigma}(u_{i})$ is zero (i.e. all reviews have the same ratings) we set $ NR(u_{i},p_{j})$ equal to zero. 
     \item  We use this normalised $NR(u_{i},p_{j})$ as a label and review text as input features (either tfidf,lda and doc2vec) to train an least square linear regressor \cite{galton1886regression} \item During prediction, regressor is used to predict a normalised review rating $PNR(u_{k},p_{m})$ for new review of user $u_{k}$ for product $p_{m}$. We recover the original user-biased score by the equation:
    \begin{equation*}
       PR(u_{k},p_{m}) = PNR(u_{k},p_{m}) * {R_{\sigma}(u_{k})} +  R_{\mu}(u_{k})  
     \end{equation*}
     Here, $PR({u_{k},p_{m}})$ and $PNR(u_{k},p_{m})$ are the predicted score and predicted normalised score (user-bias removed) respectively for user $u_{k}$  and product $p_{m}$. Since the true rating are integers, we floor or cap the $PR({u_{k},p_{m}})$ to the nearest integer in [1,5] to get the final prediction rating. This rating is used for final error calculation.
\end{enumerate}
\justify
Instead of normalising over all reviews, like previous work does, we do user specific normalization in order to implicitely identify user-specific bias. \\

\begin{figure*}[h!]
\centering
\includegraphics[scale=0.26]{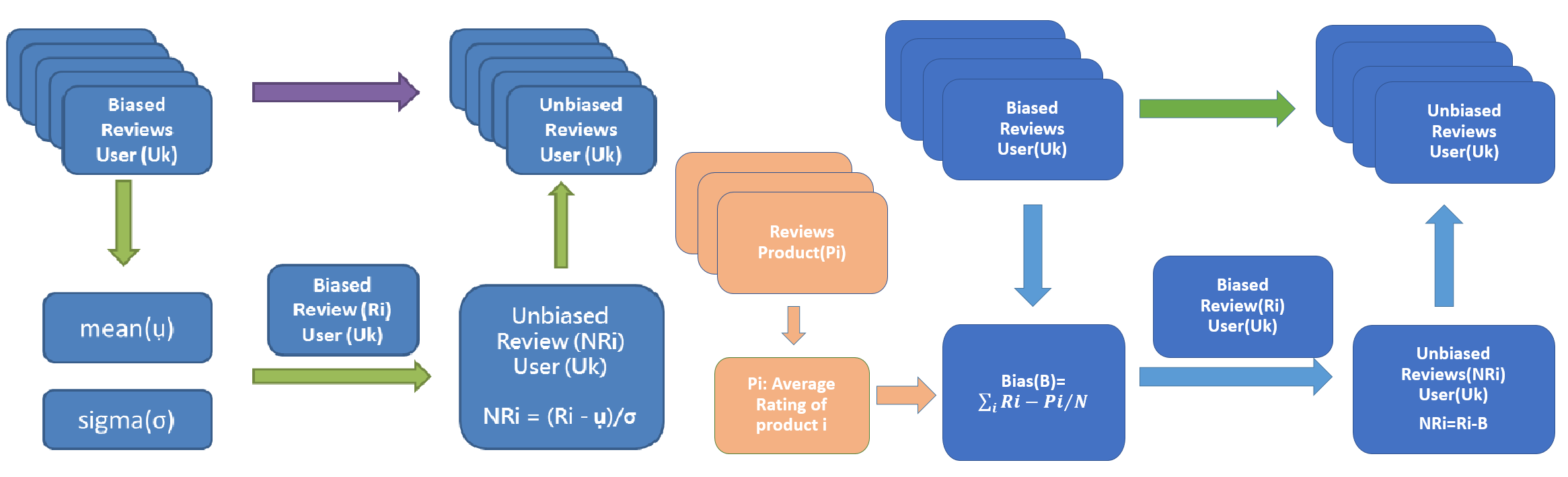}
\caption{Architecture for UBR I (L) and UBR II (R) for bias removal for specific user}
\label{removalfigure}
\end{figure*}

\subsection{User-Bias Removal-II (UBR-II)}

A product has ratings given by multiple users having positive, negative or no bias. Hence, we assume the average rating for the product is unbiased. Then, the differences of a specific user's score from this average rating can be considered that user's bias. These individual biases averaged over all products gives us the net bias for the user. This bias can then be used in a manner similar to that in Method I. The details are as follows:  
\begin{enumerate}
    \item For each product calculate the mean $R_{\mu}(p_{i})$ of the scores given by all the users. 
    \begin{equation*}
    R_{\mu}(p_{i}) = \frac{1}{N_{p_{i}}} \sum_{j=1}^{N_{p_{i}}} {R(u_{j},p_{i})}
    \end{equation*}
     Here, $N_{p_{i}}$ is the number of reviews for product $p_{i}$.
     \item For a user $u_{j}$ and product $p_{i}$, $R(u_{j},p_{i})$-$R_{\mu}(p_{i})$ is the bias of that user for product $p_{i}$. Now calculate the net bias of that user.
     $$B(u_j)= \frac{1}{N_{u_{j}}}\sum_{i=1}^{N_{u_{j}}} {R(u_{j},p_{i})-R_{\mu}(p_{i})}$$
      Here, $N_{u_{j}}$ is the number of products reviewed by user $u_{j}$. 
    \item For each review score, calculate the user-bias removed score $NR(u_{j},p_{i})$ as follows :
     \begin{equation*}
       NR(u_{j},p_{i}) = R(u_{j},p_{i})- B(u_j)
     \end{equation*}
     Here, $NR(u_{j},p_{i})$ represents the normalised score (after user-bias removal) for user $u_{j}$ and product $p_{i}$. 
     \item Using this Normalised $NR(u_{j},p_{i})$ score as labels and text as input features (either tfidf, lda and doc2vec) to train an least square linear regressor \cite{galton1886regression} . 
     
     \item During prediction, regressor model is used to predict the normalised review rating $PNR(u_{k},p_{m})$ for new review of user $u_{k}$ for product $p_{m}$. We recover the original user-biased score by the equation:
    \begin{equation*}
       PR(u_{k},p_{m}) = PNR(u_{k},p_{m}) +  B(u_k)
     \end{equation*}
     Here, $PR(u_{k},p_{m})$ and $PNR(u_{k},p_{m})$ are the predicted score and predicted normalised score (user-bias removed) respectively for user $u_{k}$  and product $p_{m}$. Since the true rating are integers , we used floor or cap $PR(u_{k},p_{m})$ to nearest integer in {1, 2, 3, 4, 5} to get the final prediction score. This will be used to calculate the final error.
\end{enumerate}
\justify
Note that, instead of normalising over all reviews, we do product specific zero mean normalization and thus consider only review scores of products that the user has reviewed to gauge his bias. In both, UBR-I and UBR-II, we assume that user has reviews at least one product. This is a fair assumption, since a new user can't provide any information or cues to model their bias.



\section{Experiments}

\subsection{Dataset Description}
We use the Amazon Food Review Dataset \cite{McAuley} consisting of 568,454 reviews by Amazon users up to October 2012. The dataset is publicly available for download from the Kaggle site (www.kaggle.com) as Amazon Fine Food Reviews \cite{Kaggle}. Each review has a ReviewId, UserId, Score, Text and a brief Summary of the review.

 We also experiment on two major categories, \textit{Electronics} and \textit{Movies and TV}, in the Amazon e-commerce dataset. These are described in detail in \cite{McAuley} and \cite{McAuley1}. We use a 4:1 train-test split on all datasets. We uniquely identify each user by thier {\em UserId} provided in the dataset. Similarly, each product is identified by the {\em ProductId} field.





Here, $P_{true}$ and $P_{est}$ are the true and predicted value respectively for test dataset and $N$ is the total number of samples used in the test set.

\subsection{Baselines}
We compare our methods to 5 statistical methods and 4 classification methods that don't model user-bias:
\begin{itemize}
\setlength\itemsep{0.1em}
    \item Majority Voting: Predict score for a review as the mode of all reviews.
    \item User Mode User Mode: Predict score for a review by user $u_{i}$ as the mean/ mode of all review scores of user $u_{i}$.
    \item Product Mean / Product Mode : Predict score for a review of product $p_{j}$ as the mean/ mode of all review scores of the product $p_{j}$.
    \item LinearSVM: Train a multi-class classifier (Linear SVM one-vs-rest ) with text features from text+summary field to predict scores (class) for a given review.
    \item NaiveNB: Train a multi-class classifier (Bernoulli/Multinomial Naive Bayes) with text features from text+summary field to predict scores (class) for a given review.
    \item Decision Tree: Train a multi-class classifier (Decision Tree) with text features from text+summary field to predict scores (class) for a given review.
\end{itemize}

Majority voting method is independent of specific product or specific user. The first five baseline methods are independent of extracted review text features. We evaluate our models on both bigram (bi) and unigram (uni) vocabulary with all baselines. All baseline give integer rating. Implementation for both the methods (UBR-I and UBR-II) and baselines is available on github \footnote{https://github.com/warahul/UBR}. 

\subsection{Results}
We compare the baselines with both our methods i.e. user-bias Removal-I (UBR-I) and user-bias Removal-II (UBR-II) using standard root mean square error (rmse) as it is a more relavant score to measure the performance of relative scoring than accuracy is. We evaluate our approach with three feature formation techniques tf-idf \cite{salton1986introduction} (25K Vocabulary), LDA \cite{Blei:2003} (ntopics = 100) and Doc2Vec (PV-DBOW) \cite{le2014distributed} to check the effect of the feature formation technique. In tf-idf we compare our approach on both unigram (25K Vocabulary) and bi-grams (25K Vocabulary). All the hyper-parameters are tuned and the performance reported is the best performance. \\


\begin{table}[h!]
\centering
\caption{RMSE results for Amazon food dataset (assume unigram vocabulary unless mention, values in red show best performance, the UBR method of this paper)}
\begin{tabular}{|c|c|c|c|} 
 \hline
 \bf Methods &\bf  tf-idf &\bf  LDA &\bf   PV-DBoW \\ [0.5ex] 
 \hline
 Majority Voting & 1.535 & 1.535 & 1.535  \\ 
 User Mean & 0.599 & 0.599 & 0.599  \\
 User Mode & 2.557 & 2.557 & 2.557\\
 Product Mean & 1.140 & 1.140  & 1.140\\
 Product Mode & 1.746  & 1.746 & 1.746\\
 LinearSVM & 0.888 & 1.494 & 1.06 \\
 LinearSVM (bi) & 0.737 & - & - \\
 MultinomialNB & 1.360& 1.535& 1.535\\
 MultinomialNB(bi) & 1.047& -&-\\
 BernoulliNB &1.173 &1.535 & 1.182\\
 Bernoulli NB(bi) & 1.041&- &-\\
 Decision Tree &1.042 & 1.259& 1.485\\
 Decision Tree (bi) &1.015 &- &-\\
 UBR-I & 0.546 & {\color{red}0.597} & {\color{red}0.56} \\
 UBR-I (bi) & {\color{red}0.529} & - & -  \\
 UBR-II & 0.669 & 0.778 & 0.71  \\
 UBR-II (bi) & 0.642 & - & -  \\
 \hline
\end{tabular}
\label{table:1}
\end{table}


\begin{table}[h!]
\centering
\caption{RMSE results for Amazon e-Commerce Electronic dataset (assume unigram vocabulary unless mention, values in red show best performance, the UBR method of this paper)}
\begin{tabular}{|c|c|c|c|} 
 \hline
 \bf Methods &\bf tf-idf &\bf LDA &\bf  PV-DBoW \\ [0.5ex] 
 \hline
 Majority Voting & 1.417  & 1.417  & 1.417\\ 
 User Mean & 1.022  & 1.022  & 1.022\\
 User Mode & 1.278  & 1.278  & 1.278\\
 Product Mean  & 1.095 & 1.095 & 1.095\\
 Product Mode & 1.358 & 1.358 & 1.358\\
 LinearSVM & 0.932 & 1.434 & 1.1\\
 LinearSVM (bi)& 0.805 & - & -\\
 MultinomialNB &1.299 & 1.417& 1.417\\
 MultinomialNB(bi) &1.045 &- &-\\
 BernoulliNB &1.225 &1.417 & 1.1706\\
 Bernoulli NB(bi) &1.137 &- &-\\
 Decision Tree &1.237 &1.434 & 1.480\\
 Decision Tree (bi) & 1.199&- &-\\
 UBR-I & 0.815 & {\color{red}0.988} & {\color{red}0.86}\\
 UBR-I (bi) & 0.763  & - & - \\
 UBR-II & 0.821 & 1.011 & 0.9 \\
 UBR-II (bi) & {\color{red}0.761} & - & -\\
 \hline
\end{tabular}
\label{table:rmse_electronics}
\end{table}


\begin{table}[h!]
\centering
\caption{RMSE results for Amazon e-Commerce Movies dataset (assume unigram vocabulary unless mention, values in red show best performance, the UBR method of this paper)}
\begin{tabular}{|c|c|c|c|} 
 \hline
 \bf Methods &\bf tf-idf &\bf LDA &\bf  PV-DBoW \\ [0.4ex] 
 \hline
 Majority Voting & 1.494 & 1.494 & 1.494\\ 
 User Mean & 1.005  & 1.005  & 1.005 \\
 User Mode & 1.258 & 1.258 & 1.258\\
 Product Mean & 1.066  & 1.066  & 1.066\\
 Product Mode & 1.347 & 1.347 & 1.347\\
 LinearSVM & 0.936 & 1.273 & 1.08 \\
 LinearSVM (bi)& 0.853 & - & - \\
 MultinomialNB &1.271 &1.494 &1.494\\
 MultinomialNB(bi) & 1.041& -&-\\
 BernoulliNB & 1.264&1.494 & 1.098\\
 Bernoulli NB(bi) &1.206 & -&-\\
 Decision Tree & 1.294&1.445 & 1.466\\
 Decision Tree(bi) & 1.270& -&-\\
 UBR-I & 0.818 & {\color{red}0.959} & {\color{red}0.87}\\
 UBR-I (bi) & 0.783 &- &-\\
 UBR-II & 0.814  & 0.982 & {\color{red}0.87} \\
 UBR-II (bi) & {\color{red}0.775} &- &-\\
 \hline
\end{tabular}
\label{table:rmse_movies}
\end{table}

Table 1 shows results for Amazon Fine Food Reviews. It is clear from Table 1 that UBR-I and UBR-II generally outperform all six baselines for all feature formation techniques (tf-idf, LDA and Doc2Vec). Tf-idf with bigram features outperforms tf-idf with unigram possibly because of automatic handling of negation bigrams in the text. We also experiment with Amazon e-Commerce electronics and movies \& TV data-sets. The corresponding results are shown in Table 2 and Table 3 respectively. Again, UBR-I and UBR-II generally outperform all six baselines for all feature formation techniques (tf-idf, LDA and Doc2Vec). Note, $'-'$ in all tables represent not applicable.

\section{Related Work}
Most relevant work that handles user-bias and is similar to our approach is described in \cite{seroussi2010collaborative}. It is based on memory based collaborative filtering for score prediction. Multiple score prediction models are trained, one for each user,  using reviews and scores corresponding to that user. Compared to our approach \cite{seroussi2010collaborative} requires multiple models, one for each user. In general, many users have very few review texts which results in poor models for user specific classifiers/regressors due to sparsity in training data. Another problem is large prediction and training time along with space requirements since it has to learn multiple models. For good generalization performance, their models have large sample, space and time complexity. Since we use a single regressor in our model the sample complexity needed is much lower. In addition, it is also fast in training and prediction. \cite{li2011incorporating} also used multiple user-product specific models corresponding to each user-product pair to handle bias. They use coordinate descent or alternate minmax to learn parameters. Similar to \cite{seroussi2010collaborative}, the method requires large sample complexity and has high space and time complexity for good generalization performance.


Other approaches mentioned in \cite{tang2015user},\cite{tang2015learning} and \cite{chen2016learning} use deep learning models to incorporate user-bias, product bias or both. All these models have a large number of training parameters i.e. weights of the deep network. Also, model complexity increases because of user and product specific parameters. Compared to other approaches, these models require large training datasets as well. Other approaches are generally described for binary i.e. 0-1 score prediction where the user-bias problem is not as germane as for ordinal rating prediction.

In real world data sets there are not enough reviews per user to train separate models for each user. In all the three amazon datasets, the distribution of number of review per user have a long tail as shown in Figure \ref{tailfigure}. Also training separate classifier over user is intractable due to large sample requirement, large training and prediction time and large storage requirements.

\section{Conclusion}
We consider the problem of user-bias in review score prediction and suggest two simple statistical approaches to reduce prediction error (RMSE). We experimented on three popular feature vector representations, tfidf, LDA and Doc2Vec on the Amazon fine food reviews dataset and on two major categories of the Amazon e-commerce dataset (\textit{Electronics} and \textit{Movies and TV}). Our approach showed improved RMSE performance as compared to baseline approaches which don't remove user-bias. Compared to other methods which use multiple classifiers for every user, our proposed methods only use a single global classifier for predicting scores. Our proposed methods have lower sample, space and time complexity compared to other methods mentioned in literature. 

\section{Future Work}
Currently, we only used review text to predict scores and not for user-bias removal. We can define bias jointly over different types of feedback (like text sentiments, review scores etc.) as a future direction. We plan to extend proposed model to take into account positive and negative terms sentiment along with scores for more accurate bias modeling. We can jointly model both,  individual user-bias(UBR-I) and collective user-bias for a given product i.e. product Bias(UBR-II) into a combined model (UBR-III).


\section*{Acknowledgments}
The authors wants to thank Nagarajan Natarajan (Post-Doc, Microsoft Research, India), Janish Jindal (Student, IIT Kanpur) and Bhargavi Paranjape (Research Fellow, Microsoft Research, India) for encouraging and valuable feedback .


\bibliography{eacl2017}
\bibliographystyle{eacl2017}


\end{document}